\documentclass{article}

\usepackage{iclr2027_conference,times}
\usepackage[utf8]{inputenc}
\usepackage[T1]{fontenc}
\usepackage{amsmath}
\usepackage{amsfonts}
\usepackage{algorithm}
\usepackage{algpseudocode}
\usepackage{graphicx}
\usepackage{wrapfig}
\usepackage{booktabs}
\usepackage{multirow}
\usepackage{nicefrac}
\usepackage{microtype}
\usepackage{xcolor}
\usepackage{enumitem}
\usepackage{hyperref}
\usepackage{url}

\usepackage{pgfplots}
\usepackage{xcolor}
\usepackage{wrapfig}
\pgfplotsset{compat=1.18}

\definecolor{gsblue}{RGB}{85,110,255}
\definecolor{oursred}{RGB}{255,95,95}

\title{Toward Comprehensive 3D Grounding: Orientation Grounding through Vision-Language Models}

\iclrfinalcopy
\author{%
\makebox[\dimexpr\textwidth-2\tabcolsep\relax][c]{%
\begin{tabular}{c}
Tuo Liang$^{1}$ \quad
Disheng Liu$^{1}$ \quad
Nengbo Wang$^{1}$ \quad
Vipin Chaudhary$^{1}$ \quad
Yu Yin$^{1}$ \\[0.5em]
\normalfont $^{1}$Case Western Reserve University
\end{tabular}%
}%
}
\begin{document}

\maketitle
\fancyhead[L]{}
\begin{abstract}
Grounding is a core capability of spatial vision-language models, yet most existing work focuses only on where a referred object is. Many 3D tasks also require knowing how it is oriented. Although existing 3D VLMs may predict oriented boxes, box pose does not explicitly capture object-centric orientation or symmetry-induced ambiguities. We introduce \emph{orientation grounding}, a referring grounding task that predicts an object's 6D orientation and axial symmetry from a language or box query in single-view or multi-view scenes. To support this task, we construct \textbf{ReferOri}, with 331K multi-view and 387K single-view orientation-grounding queries obtained through scalable reconstruction, consistency checking, and human verification. We further present \textbf{OG-VLM}, which adapts a 3D VLM with structured box/orientation outputs, sign and symmetry tokens, and geometry-aware auxiliary losses. Across single-view and multi-view benchmarks, OG-VLM substantially outperforms orientation-aware VLM baselines and surpasses object-level orientation foundation models on scene-level referring benchmarks, showing that explicit orientation grounding is a distinct and learnable capability beyond localization. Downstream results validate its benefit for orientation-related spatial reasoning.
\end{abstract}

\section{Introduction}
\label{sec:intro}
Recent progress  in spatial vision-language models (VLM), especially 3D VLM, has made grounding a central interface between language and geometry. Most existing grounding benchmarks and models focus on localization: given a referring expression, the model predicts a 2D or 3D box that identifies where the object is~\citep{hong20233d,chen2024grounded,zhang2024llava,li2025seeground,arnaud2025locate}. More recent systems further use grounded objects as intermediate representations for spatial reasoning and 3D VQA~\citep{ma2025spatialreasoner,chen2025reasoning}. \textbf{This trend makes grounding more than a supporting component; it is becoming a prerequisite for reliable spatial understanding.}

\begin{figure}[t]
  \centering
  \includegraphics[width=\linewidth]{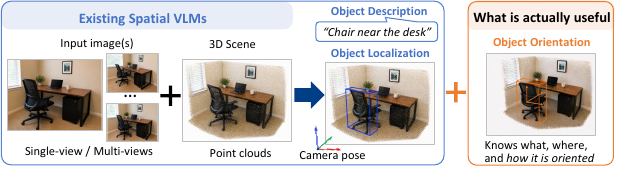}
  \vspace{-2em}
  \caption{Orientation grounding extends localization grounding from predicting where a referred object is to estimating how it is oriented. ReferOri provides single-view and multi-view annotations with object boxes, 6D orientation, and axial symmetry labels, enabling OG-VLM to ground both object position and object-centric direction.}
  \vspace{-1.5em}
  \label{fig:teaser}
\end{figure}

Localization alone, however, only specifies part of an object's state. Many spatial and scene-level tasks~\citep{linghu2024multi,ma20253dsrbench} also require object-centric orientation: as shown in Fig~\ref{fig:teaser}, model may know there is a chair in front of himself but doesn't understand the true direction he can sit down. This information is not equivalent to the pose of an enclosing box. Although recent 3D VLMs~\citep{zheng2025learning,linghu20263d} use oriented bounding-box supervision from datasets such as KITTI, SUN RGB-D, nuScenes, ARKitScenes, and EmbodiedScan~\citep{geiger2013vision,song2015sun,caesar2020nuscenes,baruch2021arkitscenes,wang2024embodiedscan}, the resulting orientation remains coupled to box geometry, coordinate conventions, and localization quality. In contrast, object-centric orientation asks how the referred object itself is directed, which often depends on part structure, local shape, symmetry and environment.

Despite the practical importance of orientation, dedicated studies on orientation grounding remain limited. Existing spatial VLMs can sometimes infer coarse directions from category priors or canonical poses~\citep{ma2022sqa3d,azuma2022scanqa,chen2024grounded}, but such behavior is fragile under non-canonical object poses, partial occlusion, and symmetric or near-symmetric shapes. Studying orientation as an explicit grounding target therefore exposes a different capability from localization: the model must both identify the referred object and recover its object-centric axes from visual and geometric evidence.

In this work, we define \emph{orientation grounding} as a natural next step beyond localization grounding toward more comprehensive 3D world understanding. Given a queried single-view or multi-view 3D scene and either a referring expression or an object box, the goal is to predict the target object's 6D orientation together with axial symmetry labels along its canonical axes. This formulation raises three challenges: referring orientation labels are scarce, symmetry introduces inherent orientation ambiguities, and autoregressive VLMs are prone to sign flips and inconsistent binary predictions in structured numeric outputs.

To address these challenges, we introduce \textbf{ReferOri}, a referring orientation-grounding dataset built by linking existing referring datasets with scalable scene and object reconstruction. ReferOri combines multi-view consistency checks with human verification to produce object orientation and axial symmetry annotations, yielding 331K multi-view and 387K single-view training queries and supporting both scene-level referring grounding and object-centric orientation estimation.
Building on ReferOri, we present \textbf{Orientation-Grounding VLM} (\textbf{OG-VLM}), an adaptation of an existing 3D VLM for orientation grounding. OG-VLM introduces explicit \texttt{<bbox>} and \texttt{<ori>} output spans, dedicated tokens for numerical signs and symmetry labels, and lightweight geometry-aware heads for box, orientation, and symmetry supervision. Experiments show substantial gains over VLM-based spatial grounding/reasoning models and object-level orientation foundation models, demonstrating that orientation grounding is a distinct and learnable capability beyond localization grounding alone.


In this paper, our contributions can be summarized as follows:
\begin{itemize}[leftmargin=*, nosep]
    \item We introduce \textit{orientation grounding} as a referring 3D grounding task that extends localization grounding to object-centric orientation and axial symmetry, and construct \textbf{ReferOri}, a scalable benchmark spanning single- and multi-view scenes through an automated data construction pipeline.


     \item We present \textbf{OG-VLM}, an adaptation of an existing 3D VLM for orientation grounding that combines structured spatial output tokens with geometry-aware losses for box, orientation, and symmetry supervision.
    
    \item We conduct extensive experiments across multi-view, single-view, and orientation-specific benchmarks, showing consistent gains over orientation-aware VLM baselines and object-level orientation foundation models, while downstream evaluations further demonstrate the benefit of orientation grounding for orientation-related spatial reasoning.
\end{itemize}

\vspace{-1em}
\section{Related Work}

\subsection{3D Referring Grounding}
3D referring grounding links language to scene geometry by localizing queried objects. Prior work develops datasets and models for predicting target boxes from reconstructed scenes or multi-view observations~\citep{chen2020scanrefer,zhang2023multi3drefer,hong20233d,chen2024grounded,zhang2024llava,li2025seeground,arnaud2025locate}, while recent spatial VLMs further use grounding as an intermediate representation for reasoning and 3D question answering~\citep{ma2025spatialreasoner,chen2025reasoning,liu2025deconstructing}. These works establish localization as a foundation for spatial understanding, but mainly predict the position and extent of referred objects. We retain this referring setting while extending the target to object-centric orientation and symmetry.

\subsection{Orientation Dataset}
Existing orientation datasets mainly fall into three categories. Object-level 3D datasets~\citep{reizenstein2021common,collins2022abo,wu2023omniobject3d,ahmadyan2021objectron} provide canonical geometry and object-centric poses, but typically lack scene context. Single-view scene-level datasets~\citep{xiang2016objectnet3d,xiang2014beyond,ma2024imagenet3d} often infer pose through CAD matching, making them dependent on external 3D priors and sensitive to occlusion, truncation, and viewpoint ambiguity. Multi-view scene-level datasets~\citep{geiger2013vision,song2015sun,caesar2020nuscenes,zhang2024omni6dpose} provide oriented boxes or pose annotations, but mainly target detection and pose estimation rather than referring orientation grounding. ARKitScenes~\citep{baruch2021arkitscenes} and EmbodiedScan~\citep{wang2024embodiedscan} offer rich scene geometry and can support referring-oriented training after adaptation. Our pipeline extends this line by constructing large-scale orientation and symmetry annotations directly under referring settings for both single-view and multi-view data.

\subsection{Orientation Estimation}
\noindent\textbf{Non-VLM Related.}
Earlier approaches~\citep{avetisyan2019scan2cad,ma2024imagenet3d,chen2024zeropose} recovered object orientation through computer-aided design (CAD) matching and alignment in images or 3D scenes, relying on large 3D CAD repositories and remaining sensitive to occlusion, truncation, and structural variation. Later methods moved toward direct pose or orientation estimation from visual input, including PoseCNN~\citep{xiang2017posecnn}, the OnePose series~\citep{sun2022onepose,he2022onepose++}, the GenPose series~\citep{zhang2023genpose,zhang2024omni6dpose}, POPE~\citep{fan2024pope}, and the Orient Anything series~\citep{wang2024orient,wang2026orient}. More recent reconstruction-based methods, such as One2Any~\citep{liu2025one2any} and SAM3D~\citep{chen2025sam}, estimate orientation by reconstructing object geometry. These works mainly formulate orientation as pose estimation, CAD alignment, or geometry recovery, whereas we study it as a grounding target under language and scene context.

\noindent\textbf{VLM-Related.}
Recent VLM-based works have begun to incorporate orientation into spatial understanding, but it is rarely treated as a first-class grounding target with dedicated representation and supervision. SpatialReasoner~\citep{ma2025spatialreasoner} uses orientation grounding to support downstream reasoning, yet inaccurate object-centric orientation can become a bottleneck for orientation-dependent reasoning, as reflected in our later evaluations. SOFAR~\citep{qi2025sofar} uses a separate semantic-orientation predictor for spatial reasoning and object manipulation, but focuses on object-level semantic directions rather than scene-level referring orientation grounding. VG-LLM~\citep{zheng2025learning} predicts orientation as part of 9D oriented boxes, coupling rotation with box geometry and Euler-angle parameterization. More generally, query-conditioned object-centric orientation and symmetry-induced ambiguities remain insufficiently modeled under a unified cross-category reference frame in cluttered scenes. We address these issues with explicit referring orientation grounding, dedicated orientation/symmetry supervision, and geometry-aware modeling.
\vspace{-1em}
\section{Methodology}
\label{sec:method}

\subsection{Task Formulation}
\label{sec:task_formulation}

Given a scene observation \(X\) and a query \(q\), orientation grounding predicts
the referred object's spatial state
\(\hat{\mathcal{A}}=(\hat{B},\hat{R},\hat{S})\),
where \(\hat{B}\in\mathbb{R}^{6}\) is the predicted 3D bounding box,
\(\hat{R}\in SO(3)\) is the predicted object-centric rotation in the scene/world
coordinate system, and \(\hat{S}\in\{0,1\}^{3}\) denotes the predicted axial
symmetry along the canonical \(x\), \(y\), and \(z\) axes.
We use \(\mathcal{A}=(B,R,S)\) to denote the corresponding ground-truth
annotation throughout the paper.
The query can be a natural-language referring expression or a box query.
The model predicts rotation using the continuous 6D representation
of \(R\)~\citep{zhou2019continuity}, i.e., the first two axes of the rotation
basis, which is converted to \(\hat{R}\in SO(3)\). Symmetry matters because several orientations can be perceptually equivalent
for symmetric objects. We therefore define the symmetry-aware discrepancy
\[
d_{\hat S}(R,\hat R)
=
\min_{Q\in\mathcal{G}(\hat S)}
\arccos
\left(
\frac{
\mathrm{tr}\!\left(\hat R(RQ)^\top\right)-1
}{2}
\right),
\]

\begin{wrapfigure}{r}{0.5\textwidth}
\vspace{-1em} 
\begin{minipage}{\linewidth}
\begin{algorithm}[H]
\caption{Orientation annotation pipeline}
\begin{algorithmic}[1]

\Require Referred sample $(I,b)$ or multi-view sample $\{(I_t,b_t)\}_{t=1}^T$
\Ensure Orientation grounding annotation $\mathcal{A}=(B,R,S)$

\If{single-view}
    \State Estimate $T_{\text{world}\leftarrow\text{cam}}$ with Map Anything
    \State Obtain object mask $M$ from $b$ using SAM2
    \State Reconstruct object point cloud $\mathcal{P}$ and pose $T_{\text{cam}\leftarrow\text{obj}}$ with SAM3D
    \State Estimate symmetry labels $S$ from $\mathcal{P}$
    \State Compute world-frame orientation $R$
    \State Construct 3D box $B$
\Else
    \State Select the top-2 most informative views
    \State Estimate two candidate world-frame orientations $\{R_1,R_2\}$ using the same SAM2--SAM3D pipeline
    \If{$\angle(R_1,R_2)<\tau$}
        \State Set $R \leftarrow R_1$
    \Else
        \State Apply lightweight human correction on a coarse annotation to obtain $R$
    \EndIf
    \State Estimate $S$ and construct $B$
\EndIf

\State \Return $\mathcal{A}=(B,R,S)$

\end{algorithmic}
\label{alg:annotation}
\end{algorithm}
\end{minipage}
\vspace{-40pt} 
\end{wrapfigure}

When no symmetry ambiguity is present, $\mathcal{G}(\hat S)=\{I\}$.
This formulation separates orientation errors from equivalent poses induced by
object symmetry. We apply the same symmetry-aware treatment at inference time.

\subsection{Data Construction}
\label{sec:data_construction}

\begin{wrapfigure}{r}{0.48\columnwidth}
    \vspace{-1.2em}
    \centering
    \includegraphics[width=\linewidth]{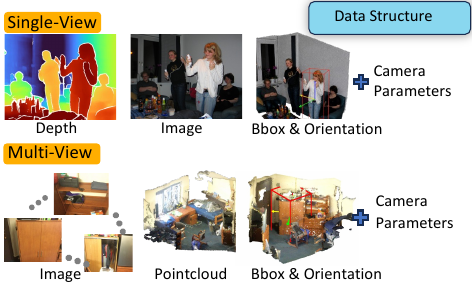}
    \vspace{-2em}
    \caption{Overview of ReferOri dataset structure.}
    \label{fig:dataset_structure}
    \vspace{-1em}
\end{wrapfigure}


\textbf{ReferOri} attaches object-centric orientation and symmetry annotations
to existing referring data. For \textbf{single-view} scenes, we use
RefCOCO~\citep{yu2016modeling}. For \textbf{multi-view} scenes, we annotate
ScanNet~\citep{dai2017scannet} objects and transfer the resulting annotations to
ScanNet-based referring datasets for training and evaluation.
Following the notation in Sec.~\ref{sec:task_formulation}, each referred object
is annotated as \(\mathcal{A}=(B,R,S)\), where \(B\) is the 3D box,
\(R\in SO(3)\) is the world-frame object orientation, and
\(S\in\{0,1\}^{3}\) denotes axial symmetry.
Algorithm~\ref{alg:annotation} summarizes the overall annotation procedure,
with the single-view and multi-view branches detailed below.


\noindent\textbf{Single-view construction.}
Given an image \(I\) and a referred bounding box \(b\), we first use Map Anything~\citep{keetha2025mapanything} to estimate scene depth and camera parameters, yielding a camera-to-world transformation \(T_{\text{world}\leftarrow\text{cam}}\). We then apply SAM2~\citep{ravi2024sam} to refine the target mask \(M\), and feed \((I,M)\) into SAM3D~\citep{chen2025sam} to reconstruct an object point cloud \(\mathcal{P}\) together with an object-to-camera transformation \(T_{\text{cam}\leftarrow\text{obj}}\). Based on \(\mathcal{P}\), we estimate the axial symmetry labels \(S\), and treat the canonical coordinate system of the reconstructed object as an orientation proxy. The final world-frame orientation is computed as $R_{\text{obj}}^{\text{world}} = R_{\text{world}\leftarrow\text{cam}} \, R_{\text{cam}\leftarrow\text{obj}}$.

\noindent\textbf{Multi-view construction.}
For ScanNet scenes, we select the two most informative views of the referred object, approximated by the largest visible mask area. We estimate object orientation independently from both views using the same SAM2--SAM3D pipeline and transform the results into the shared world frame using the native multi-view camera parameters. If the angular discrepancy between the two estimates is below a threshold \(\tau\), we directly accept the annotation; otherwise, we retain one estimate as a coarse annotation and apply lightweight human correction.


\noindent\textbf{Verification.}
We validate annotation quality through human inspection.
For single-view data, five rounds of 100 randomly sampled annotations are
reviewed for referent correctness, orientation consistency with the source
image and reconstructed geometry, and coordinate consistency, yielding a
96\% human-verified correctness rate.
For multi-view ScanNet data, all $\sim$10K object annotations are manually
inspected and corrected when necessary.
We additionally audit 800 images with two independent evaluators, obtaining
97.5\% inter-annotator agreement.
Detailed verification protocols and annotation-module validation are provided
in Appendix~\ref{app:human_verification}. The resulting structure of ReferOri after annotation and verification is illustrated in Fig.~\ref{fig:dataset_structure}.

\begin{figure*}
\vspace{-1em}
    \centering
    \includegraphics[width=0.9\linewidth]{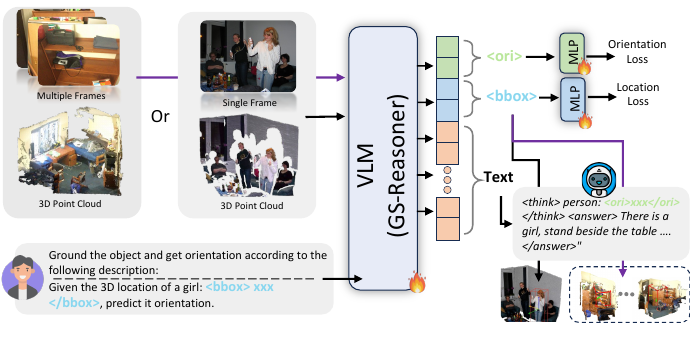}
    \vspace{-3em}
    \caption{Overview of OG-VLM. Given multi-view frames, a single frame, or a 3D point cloud together with a text or box query, the VLM backbone generates structured grounding outputs with separate \texttt{<bbox>} and \texttt{<ori>} spans. Lightweight heads attached to the corresponding tokens provide localization, orientation, and symmetry supervision through geometry-aware losses.}
    \vspace{-1em}
    \label{fig:pipeline}
\end{figure*}

\subsection{OG-VLM Overview}
\label{sec:model_overview}
OG-VLM predicts the 3D box, orientation, and axial symmetry of a queried object
(Fig.~\ref{fig:pipeline}) from single-/multi-view input or a point cloud, together
with a language or box query.
Built on GS-Reasoner~\citep{chen2025reasoning}, it introduces two extensions:
separate \texttt{<bbox>} and \texttt{<ori>} spans with dedicated sign and symmetry
tokens (Sec.~\ref{sec:special_token}), and lightweight heads on span-closing hidden
states for geometry-aware box, orientation, and symmetry supervision
(Sec.~\ref{sec:geo_loss}).

\subsubsection{Special Token Design for Orientation Grounding}
\label{sec:special_token}
GS-Reasoner represents 3D boxes as plain-text coordinates inside a
\texttt{<bbox>} span.
Extending this format to orientation introduces three issues: interference
between box and orientation values, severe errors from sign flips in rotation
components, and inconsistent free-form symmetry labels.
Following prior LLM-based grounding models~\citep{peng2023kosmos,chen2024grounded},
we address these issues with separate structured spans and dedicated sign and
symmetry tokens.

To improve the stability of structured spatial prediction, we explicitly disentangle localization and orientation outputs using
\(Y=[\,y^{\text{bbox}},\,y^{\text{ori}}\,]\),
where \(y^{\text{bbox}}=\texttt{<bbox>}(\tilde{\mathbf b})\texttt{</bbox>}\), and \(y^{\text{ori}}=\texttt{<ori>}(\tilde{\mathbf o},\mathbf s)\texttt{</ori>}\).
Here, \(\mathbf b\in\mathbb{R}^6\) denotes the 3D bounding box, \(R\in SO(3)\) denotes the object orientation, \(\mathbf o\in\mathbb{R}^6\) is the continuous 6D representation of \(R\) used in training, and \(\mathbf s\in\{0,1\}^3\) denotes the axial symmetry labels.

Rather than generating signed values as free-form text, we factorize each scalar $v$ of $\mathbf b$ and $\mathbf o$ into a sign token and its magnitude, and we write each symmetry label as a dedicated token:
\begin{equation}
v\mapsto\big(\sigma(v),\,|v|\big),\quad
\sigma(v)=
\begin{cases}
\texttt{<+>}, & v\ge 0,\\
\texttt{<->}, & v<0,
\end{cases}
\qquad
s_a\mapsto
\begin{cases}
\texttt{<s\_yes>}, & s_a=1,\\
\texttt{<s\_no>}, & s_a=0.
\end{cases}
\label{eq:value_tokens}
\vspace{-1em}
\end{equation}

This design reduces interference between box and orientation prediction while better matching token-based language modeling. Explicit sign tokens separate coarse spatial direction from numeric magnitude, reducing severe errors caused by sign flips. Discrete symmetry tokens likewise avoid unstable free-form binary generation and improve output consistency.



\subsubsection{Geometry-Aware Loss for Grounding}
\label{sec:geo_loss}
The language-modeling loss treats each numeric token as an independent class
and therefore does not explicitly reflect geometric distance.
For example, predictions with substantially different coordinate errors may
receive similar token-level penalties when they differ from the target by a
similar number of tokens.
Moreover, the language-modeling objective alone does not enforce geometric
constraints, such as orthogonality between the predicted orientation axes.
To provide explicit geometric supervision, we attach separate lightweight MLP
heads to the hidden states of the corresponding tokens for box,
orientation, and symmetry prediction.
These heads provide task-specific supervision for each spatial component (\textit{i.e.,} $\mathcal{L}_\text{box}$, $\mathcal{L}_\text{ori}$, $\mathcal{L}_\text{sym}$) and
augment the language-modeling objective:
\begin{equation}
\mathcal{L}
=
\mathcal{L}_{\text{LM}}
+
\lambda_{\text{box}}\mathcal{L}_{\text{box}}
+
\lambda_{\text{ori}}\mathcal{L}_{\text{ori}}
+
\lambda_{\text{sym}}\mathcal{L}_{\text{sym}}.
\label{eq:total_loss}
\end{equation}

\noindent\textbf{Box loss.}
From the hidden state at the \texttt{</bbox>} token, we regress a 3D box
\(B=[x_{\min},y_{\min},z_{\min},x_{\max},y_{\max},z_{\max}] \in \mathbb{R}^6\),
and define
\begin{equation}
\mathcal{L}_{\text{box}}
=
\alpha_{\text{giou}}\mathcal{L}_{\text{GIoU}}
+
\alpha_{\text{scale}}\mathcal{L}_{\text{scale}}
+
\alpha_{\text{center}}\mathcal{L}_{\text{center}},
\label{eq:box_loss}
\vspace{-0.3em}
\end{equation}
where \(\mathcal{L}_{\text{GIoU}}\), \(\mathcal{L}_{\text{scale}}\), and \(\mathcal{L}_{\text{center}}\) enforce overlap-aware alignment, relative size consistency, and normalized center consistency, respectively.

\noindent\textbf{Orientation loss.}
The ground-truth orientation lies in \(SO(3)\), while we predict its continuous 6D representation as two 3D axes \(\hat{\mathbf r},\hat{\mathbf f}\). Let \(\mathbf r,\mathbf f\) denote the ground-truth right and front axes. We define
\begin{equation}
\mathcal{L}_{\text{ori}}
=
\beta_{\text{align}}\mathcal{L}_{\text{align}}
+
\beta_{\text{ortho}}\mathcal{L}_{\text{ortho}},
\label{eq:ori_loss}
\vspace{-1em}
\end{equation}
where
\vspace{-0.5em}
\begin{equation}
\mathcal{L}_{\text{align}}
=
\frac{1}{2}
\left[
\big(1-\cos(\hat{\mathbf r},\mathbf r)+\epsilon\big)^{\lambda}
+
\big(1-\cos(\hat{\mathbf f},\mathbf f)+\epsilon\big)^{\lambda}
\right],
\qquad
\mathcal{L}_{\text{ortho}}
=
|\cos(\hat{\mathbf r},\hat{\mathbf f})|.
\label{eq:ori_terms}
\vspace{-0.5em}
\end{equation}
The two terms encourage axis alignment and mutual orthogonality, respectively. The adaptive exponent \(\lambda\) preserves sensitivity in the small-error regime. At inference, Gram--Schmidt orthogonalization converts the predicted axes into a valid rotation matrix.

\noindent\textbf{Symmetry loss.}
We further apply a cosine-based symmetry regularizer,
\begin{equation}
\mathcal{L}_{\text{sym}} = 1-\cos(\hat{\mathbf s},\mathbf s),
\label{eq:sym_loss}
\vspace{-0.5em}
\end{equation}
where \(\hat{\mathbf s}\) and \(\mathbf s\) denote the predicted and ground-truth symmetry vectors, respectively.

Overall, the proposed loss supplements token-level supervision with explicit geometric constraints, improving box consistency, axis alignment, and symmetry-aware grounding. Detailed formulations of \(\mathcal{L}_{\text{GIoU}}\), \(\mathcal{L}_{\text{scale}}\), \(\mathcal{L}_{\text{center}}\), as well as the adaptive schedule of \(\lambda\), are provided in Appendix~\ref{app:geo_loss_details}.

\vspace{-0.5em}
\section{Experiments}


\subsection{Experimental Settings}


\noindent\textbf{Training Settings.}
We use the proposed ReferOri as training data, including \textit{387K single-view} and \textit{331K multi-view} queries. 
We consider three progressively enhanced settings: \textbf{Only data}, which fine-tunes the base model on our constructed orientation-grounding data; \textbf{+ Special token}, which introduces our special-token representation; and \textbf{+ Geometric loss}, which further incorporates our geometry-aware loss. Further details on dataset construction, hyperparameter selection, and implementation can be found in Appendix~\ref{app:geo_loss_details}.

\noindent\textbf{Benchmarks and Metrics.}
We evaluate on both single-view and multi-view orientation-grounding benchmarks.
The single-view benchmarks include RefCOCO$^{*}$, RefCOCO+$^{*}$,
RefCOCOg$^{*}$~\citep{yu2016modeling}, and Ori-COCO~\citep{wang2024orient}, while the multi-view
benchmarks include ScanRefer$^{*}$~\citep{chen2020scanrefer}, Multi3DRefer$^{*}$~\citep{zhang2023multi3drefer}, and
ARKitScenes~\citep{baruch2021arkitscenes}.
Here, ($*$) denotes benchmarks whose orientation annotations are generated using our annotation pipeline.
In total, five benchmarks are annotated by our pipeline, whereas OriCOCO and ARKitScenes provide native orientation annotations. Since neither dataset relies on our annotation process, we use them exclusively for out-of-distribution (OOD) and zero-shot evaluation. 

For continuous-orientation benchmarks, we report \textbf{Acc@15} and \textbf{Acc@30}, \textit{i.e.,} the percentage of predictions whose orientation error is within \(15^\circ\) and \(30^\circ\), respectively.
For fair comparison, all methods' native orientation representations are converted to rotation matrices in a shared right-handed reference frame and scored using the same symmetry-aware \(SO(3)\) geodesic metric.
Ori-COCO is evaluated using its native eight-sector classification accuracy because it only provides discrete orientation labels.
Per-baseline conversion details are provided in
Appendix~\ref{app:eval_protocol}.

\subsection{Main Results}
\begin{table*}[t]
\caption{Results on multi- and single-view scenes. Qualitative results appear in Fig.~\ref{fig:qualitative_results} in the appendix.}
\label{tab:orientation_results}

\begin{center}
\Large
\resizebox{\textwidth}{!}{
\begin{tabular}{lccccccccccccc}
\toprule
\multirow{2}{*}{\textbf{Models}}
& \multicolumn{6}{c}{\textbf{Multi-View Scene}}
& \multicolumn{7}{c}{\textbf{Single-View Scene}} \\
\cmidrule(lr){2-7}
\cmidrule(lr){8-14}
& \multicolumn{2}{c}{\textbf{ARKitScenes}} 
& \multicolumn{2}{c}{\textbf{ScanRefer$^{*}$}} 
& \multicolumn{2}{c}{\textbf{Multi3DRefer$^{*}$}} 
& \multicolumn{1}{c}{\textbf{Ori-COCO}} 
& \multicolumn{2}{c}{\textbf{RefCOCO$^{*}$}} 
& \multicolumn{2}{c}{\textbf{RefCOCO+$^{*}$}} 
& \multicolumn{2}{c}{\textbf{RefCOCOg$^{*}$}} \\
\cmidrule(lr){2-3}
\cmidrule(lr){4-5}
\cmidrule(lr){6-7}
\cmidrule(lr){8-8}
\cmidrule(lr){9-10}
\cmidrule(lr){11-12}
\cmidrule(lr){13-14}
& Acc15 $\uparrow$ & Acc30 $\uparrow$
& Acc15 $\uparrow$ & Acc30 $\uparrow$
& Acc15 $\uparrow$ & Acc30 $\uparrow$
& Acc $\uparrow$
& Acc15 $\uparrow$ & Acc30 $\uparrow$
& Acc15 $\uparrow$ & Acc30 $\uparrow$
& Acc15 $\uparrow$ & Acc30 $\uparrow$ \\
\midrule
Spatial Reasoner 
& -- & -- 
& -- & -- 
& -- & -- 
& 14.9 
& 4.5 & 14.5 
& 4.5 & 14.0 
& 3.9 & 13.1 \\

VG-LLM 
& 10.1 & 17.4 
& 17.8 & 26.2 
& 21.5 & 33.5 
& -- 
& -- & -- 
& -- & -- 
& -- & -- \\

Ours (only data) 
& 43.1 & 45.4 
& 63.1 & 66.6
& 68.1 & 70.6 
& 60.1 
& 31.1 & 60.9
& 31.1 & 60.8 
& 28.8 & 57.8 \\

Ours (+special token) 
& 48.2 & 50.5 
& 65.2 & 68.0
& 70.5 & 73.1
& 62.2 
& 31.6 & 63.3 
& 31.7 & 61.5 
& 29.7 & 59.1 \\

Ours (+geometric loss) 
& \textbf{50.5} & \textbf{51.6} 
& \textbf{66.1} & \textbf{70.1} 
& \textbf{71.5} & \textbf{74.2} 
& \textbf{64.8} 
& \textbf{32.0} & \textbf{63.9} 
& \textbf{32.2} & \textbf{62.1} 
& \textbf{30.2} & \textbf{60.5} \\
\bottomrule
\end{tabular}
}
\end{center}

\vspace{-1em}
\end{table*}

As shown in Table~\ref{tab:orientation_results}, current spatial VLMs remain largely dominated by localization-oriented grounding, although grounding has become increasingly central to spatial understanding and reasoning. Existing baselines with partial orientation capability, including SpatialReasoner~\citep{ma2025spatialreasoner} and VG-LLM~\citep{zheng2025learning}, perform poorly on benchmarks despite being designed for single-view reasoning and multi-view oriented-box prediction, respectively. This demonstrates our hypothesis that advances in localization grounding and grounding-guided reasoning do not automatically translate into reliable object-centric orientation grounding. Predicting oriented boxes or relying on implicit orientation cues is insufficient to recover how an object itself is oriented. Moreover, representing orientation solely with three Euler angles is limited by convention dependence, discontinuities, and axis coupling, which can hinder stable learning and accurate grounding.

In contrast, our method consistently achieves large gains across both single-view and multi-view benchmarks, indicating that orientation grounding requires explicit task formulation, dedicated supervision, and tailored modeling. 
These results support our central claim that orientation grounding is a natural step beyond localization grounding toward deeper 3D world understanding. 
Localization results (in appendix Table~\ref{tab:acc25_results}) further show that our designs also improve 3D bounding box grounding.

\subsection{Comparison with Orientation Foundation Model}
\begin{wraptable}{r}{0.72\textwidth}
\vspace{-2.0em}
\centering
\caption{Comparison with orientation foundation models.}
\label{tab:orientation_specific}
\vspace{0.3em}

\resizebox{0.72\textwidth}{!}{
\begin{tabular}{lcccccc}
\toprule
\textbf{Models} 
& \multicolumn{2}{c}{\textbf{ARKitScenes}} 
& \multicolumn{2}{c}{\textbf{ScanRefer$^{*}$}} 
& \multicolumn{2}{c}{\textbf{Multi3DRefer$^{*}$}} \\
\cmidrule(lr){2-3}
\cmidrule(lr){4-5}
\cmidrule(lr){6-7}
& \textbf{Acc@15} & \textbf{Acc@30} 
& \textbf{Acc@15} & \textbf{Acc@30} 
& \textbf{Acc@15} & \textbf{Acc@30} \\
\midrule
\multicolumn{7}{l}{\textbf{Orientation foundation models}} \\
Orient Anything
& 18.3 & 35.8
& 17.4 & 34.5
& 16.9 & 33.1 \\
Orient Anything V2
& 19.8 & 43.3
& 15.8 & 37.4
& 16.9 & 35.6 \\
\midrule
Ours (+geometric loss)
& \textbf{50.5} & \textbf{51.6}
& \textbf{66.1} & \textbf{70.1}
& \textbf{71.5} & \textbf{74.2} \\
\bottomrule
\end{tabular}
}

\vspace{-1em}
\end{wraptable}
We further compare our OG-VLM with recent orientation foundation models (Table~\ref{tab:orientation_specific}). Unlike OG-VLM, which takes a queried multi-view 3D scene as input and performs orientation grounding in a scene-centric setting, existing orientation foundation models~\citep{wang2024orient, wang2026orient} assume access to a cropped target object for object-centric orientation estimation. Since object localization is assumed to be given, the latter setting is substantially less challenging and may require additional detection or segmentation modules in practice.

Nevertheless, OG-VLM achieves superior performance across both in-domain and OOD benchmarks. Notably, although trained exclusively on ScanNet-based multi-view data, it generalizes well to ARKitScenes, indicating that it learns transferable orientation representations rather than dataset-specific coordinate conventions. We attribute this advantage to OG-VLM's explicit grounding supervision and scene-aware reasoning, which enable it to jointly localize the queried target and infer its orientation from multi-view observations and surrounding scene context.

\vspace{ -1em}
\section{Analysis}

\subsection{Axis-Wise Error Analysis} 
\label{section5_1}
\begin{wraptable}{r}{0.46\textwidth}
\vspace{-2.2em}
\centering
\caption{Axis-wise mean orientation error (in degrees). X, Y, and Z denote the right, front, and up axes, respectively.}
\label{tab:axis_error_analysis}
\vspace{0.3em}

\resizebox{0.46\textwidth}{!}{
\begin{tabular}{llccc}
\toprule
\textbf{Dataset} & \textbf{Ablation} & \textbf{X} ($\downarrow$) & \textbf{Y} ($\downarrow$) & \textbf{Z} ($\downarrow$) \\
\midrule
\multirow{3}{*}{RefCOCO+$^{*}$}
& only data & 37.9 & 40.8 & 17.8 \\
& special token & 35.3 & 38.3 & 17.6 \\
& geometric loss & \textbf{35.1} & \textbf{38.0} & \textbf{17.3} \\
\midrule
\multirow{3}{*}{ScanRefer$^{*}$}
& only data & 38.4 & 38.6 & 0.1 \\
& special token & 38.1 & 38.2 & 0.1 \\
& geometric loss & \textbf{37.1} & \textbf{37.2} & \textbf{0.1} \\
\bottomrule
\end{tabular}
}

\vspace{-1em}
\end{wraptable}

As shown in Table~\ref{tab:orientation_results}, progressively adding our designs yields consistent gains, with the combination of special tokens and geometric loss performing best overall. To identify where these gains arise, we analyze axis-wise mean orientation error on RefCOCO+$^{*}$ and ScanRefer$^{*}$ in Table~\ref{tab:axis_error_analysis}. Improvements mainly occur on the X and Y axes, corresponding to the object’s right and front directions, which are also the more challenging components of orientation grounding. In contrast, Z-axis error is substantially smaller and varies little across settings, especially on ScanRefer$^{*}$. This is expected because most real-scene objects are approximately upright, making the up axis strongly constrained by gravity. Thus, our method mainly improves the harder object-centric horizontal directions, while the vertical axis is already relatively easy. The consistent X/Y error reduction from \textit{only data} to \textit{special token} and then \textit{geometric loss} further shows that both structured output design and geometry-aware supervision improve fine-grained orientation alignment rather than merely benefiting trivial axes.

\subsection{What Affects Orientation Grounding?}

Beyond the main benchmark results, we further analyze several factors that
influence orientation grounding, including input depth, object symmetry, and
localization quality. These analyses help characterize how visual and geometric
evidence, object properties, and localization accuracy affect orientation
prediction. We additionally study the effect of the number of input views in
Appendix~\ref{app:views}.

\begin{wraptable}{r}{0.7\textwidth}
\vspace{-2em}
\centering
\caption{Effect of depth on RefCOCO+$^{*}$. For localization, we report the accuracy at IoU thresholds of 25\% and 50\%. For orientation grounding, we report the accuracy of predictions whose orientation error falls within \(15^\circ\) and \(30^\circ\), as well as the mean orientation error.}
\label{tab:depth_effect_refcoco_plus}
\vspace{0.3em}

\resizebox{0.7\textwidth}{!}{
\begin{tabular}{llccccc}
\toprule
\multirow{2}{*}{\textbf{Model}}
& \multirow{2}{*}{\textbf{Setting}}
& \multicolumn{2}{c}{\textbf{Localization}}
& \multicolumn{3}{c}{\textbf{Orientation}} \\
\cmidrule(lr){3-4}
\cmidrule(lr){5-7}
&
& \textbf{Acc 25\%}
& \textbf{Acc 50\%}
& \textbf{Mean Err.\ ($^\circ$)}
& \textbf{Acc@15}
& \textbf{Acc@30} \\
\midrule

\multirow{2}{*}{only data}
& \textbf{w depth} & 68.7 & 36.3 & 29.6 & 31.1 & 60.8 \\
& \textbf{w/o depth} & 1.2 & 0.0 & 34.3 & 24.7 & 52.3 \\

\midrule
\multirow{2}{*}{special token}
& \textbf{w depth} & 68.2 & 35.6 & 29.5 & 30.9 & 61.5 \\
& \textbf{w/o depth} & 0.8 & 0.0 & 35.0 & 23.6 & 52.1 \\

\midrule
\multirow{2}{*}{geometric loss}
& \textbf{w depth} & 67.8 & 35.9 & 30.5 & 29.0 & 58.7 \\
& \textbf{w/o depth} & 2.5 & 0.1 & 33.4 & 24.4 & 53.7 \\

\bottomrule
\end{tabular}
}

\end{wraptable}






\noindent\textbf{Depth.}
We conduct the depth ablation in the single-view setting to isolate its effect from multi-view geometric consistency. As shown in Table~\ref{tab:depth_effect_refcoco_plus}, removing depth causes a dramatic drop in 3D box grounding (\textit{i.e.,} localization) but a smaller degradation in orientation grounding. This suggests that depth is crucial for metric properties such as object scale, extent, and 3D location, whereas orientation can still be partially inferred from appearance, local shape, and semantic priors. Nevertheless, the consistent drops in Acc@15 and Acc@30 show that depth remains important for precise orientation grounding. Overall, 3D box grounding relies more heavily on metric reconstruction, while orientation can additionally exploit visual and semantic cues, which also helps explain why VLMs can acquire partial orientation understanding from image-language supervision.




\begin{wraptable}{r}{0.60\textwidth}
\vspace{-2em}
\centering
\caption{Effect of object symmetry on orientation grounding.}
\label{tab:symmetry_analysis}
\vspace{0.3em}

\resizebox{0.60\textwidth}{!}{
\begin{tabular}{llcccc}
\toprule
\textbf{Dataset} & \textbf{Model} 
& \multicolumn{2}{c}{\textbf{Symmetric}} 
& \multicolumn{2}{c}{\textbf{Non-symmetric}} \\
\cmidrule(lr){3-4}
\cmidrule(lr){5-6}

& & \textbf{Acc@15} & \textbf{Acc@30}
& \textbf{Acc@15} & \textbf{Acc@30} \\
\midrule

\multirow{3}{*}{RefCOCO+$^{*}$}
& only data & 24.41 & 53.76 & 34.41 & 66.56 \\
& special token & 26.76 & 53.05 & 34.44 & 67.36 \\
& geometric loss & 28.87 & 58.69 & 35.31 & 68.79 \\

\midrule

\multirow{3}{*}{ScanRefer$^{*}$}
& only data & 64.23 & 66.37 & 59.54 & 64.31 \\
& special token & 65.97 & 68.40 & 58.37 & 63.69 \\
& geometric loss & 66.77 & 69.03 & 60.50 & 63.80 \\

\bottomrule
\end{tabular}
}

\vspace{-0.4em}
\end{wraptable}



















\noindent\textbf{Object symmetry.}
Symmetry directly determines whether multiple orientations should be treated as equivalent. We therefore compare \textit{symmetric} and \textit{non-symmetric} objects across datasets. As shown in Table~\ref{tab:symmetry_analysis}, RefCOCO+ exhibits lower accuracy on symmetric objects, suggesting greater orientation ambiguity under limited single-view geometry. In contrast, symmetric objects achieve slightly higher accuracy on ScanRefer across all settings. We hypothesize that richer multi-view geometry reduces viewpoint ambiguity, while symmetry-aware labels admit a larger set of equivalent orientations. Consequently, symmetric objects benefit more from multi-view reasoning, whereas asymmetric objects remain more sensitive to local structure, occlusion, and grounding noise.

\noindent\textbf{Effect of Symmetry and Localization Constraints.}
To disentangle genuine orientation-grounding ability from the effects of
symmetry handling and localization errors, we compare the same model under
three evaluation settings that progressively introduce symmetry adjustment
and localization constraints. As shown in Table~\ref{tab:metric_decomposition}, removing symmetry adjustment
lowers Acc@15 by 2.7 points on ScanRefer$^{*}$ and 7.2 points on
Multi3DRefer$^{*}$, but by less than 0.2 points on the RefCOCO family.
The relatively modest gains indicate that the reported orientation performance
is not primarily driven by symmetry adjustment; the model retains substantial
orientation-grounding capability without it, while explicit symmetry modeling
remains useful for intrinsically ambiguous orientations. Requiring joint
correctness with localization causes a much larger drop, particularly on
ScanRefer$^{*}$ and Multi3DRefer$^{*}$, because localization and orientation
must succeed simultaneously and errors from either component accumulate.
This distinction helps isolate orientation quality from localization failures
during evaluation. We therefore treat the joint metric as a diagnostic
complement to orientation-only evaluation, while the remaining gap suggests
that jointly acquiring multiple grounding capabilities remains challenging for
current spatial VLMs.
\begin{table}[t]
\caption{
Comparison of three evaluation settings for OG-VLM.
\textit{w/o symmetry-adjusted} directly computes orientation accuracy without
symmetry handling;
\textit{w/ symmetry-adjusted} accounts for symmetry-equivalent
orientations when computing Acc@15 and Acc@30;
and \textit{Joint} additionally requires the predicted box to satisfy a
localization IoU threshold of 0.25.
}
\label{tab:metric_decomposition}
\vspace{-2mm}
\begin{center}
\resizebox{\textwidth}{!}{
\begin{tabular}{lcccccccccc}
\toprule
\multirow{2}{*}{}
& \multicolumn{2}{c}{\textbf{ScanRefer$^{*}$}}
& \multicolumn{2}{c}{\textbf{Multi3DRefer$^{*}$}}
& \multicolumn{2}{c}{\textbf{RefCOCO$^{*}$}}
& \multicolumn{2}{c}{\textbf{RefCOCO+$^{*}$}}
& \multicolumn{2}{c}{\textbf{RefCOCOg$^{*}$}} \\
\cmidrule(lr){2-3}
\cmidrule(lr){4-5}
\cmidrule(lr){6-7}
\cmidrule(lr){8-9}
\cmidrule(lr){10-11}

& Acc@15 & Acc@30
& Acc@15 & Acc@30
& Acc@15 & Acc@30
& Acc@15 & Acc@30
& Acc@15 & Acc@30 \\
\midrule

w/o symmetry-adjusted
& 63.4 & 66.9
& 64.3 & 71.8
& 31.9 & 63.8
& 32.1 & 61.1
& 30.1 & 55.4 \\

w/ symmetry-adjusted
& 66.1 & 70.1
& 71.5 & 74.2
& 32.0 & 63.9
& 32.2 & 62.1
& 30.2 & 60.5 \\

Joint w/ IoU$\ge$0.25
& 45.2 & 47.8
& 43.0 & 45.7
& 25.8 & 50.2
& 24.5 & 47.7
& 21.5 & 39.1 \\



\bottomrule
\end{tabular}
}
\end{center}
\end{table}

\subsection{Does Orientation Grounding Benefit Downstream Spatial Reasoning?}
\label{sec:downstream_reasoning}

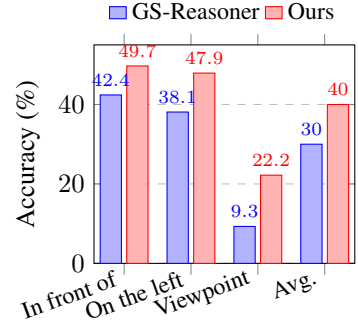
\begin{wrapfigure}{r}{0.36\textwidth}
\vspace{-2em}
\centering
\begin{tikzpicture}
\begin{axis}[
    width=0.36\textwidth,
    height=0.32\textwidth,
    ybar,
    bar width=8pt,
    ymin=0,
    ymax=55,
    ylabel={Accuracy (\%)},
    symbolic x coords={In front of, On the left, Viewpoint, Avg.},
    xtick=data,
    xticklabel style={font=\small, rotate=20, anchor=east},
    ymajorgrids=true,
    grid style=dashed,
    legend style={
        font=\small,
        at={(0.5,1.05)},
        anchor=south,
        legend columns=2,
        draw=none
    },
    legend image code/.code={
    \draw[#1] (0cm,-0.08cm) rectangle (0.22cm,0.08cm);
},
    nodes near coords,
    nodes near coords style={font=\scriptsize},
    enlarge x limits=0.15
]
\addplot coordinates {
    (In front of,42.4)
    (On the left,38.1)
    (Viewpoint,9.3)
    (Avg.,30.0)
};
\addplot coordinates {
    (In front of,49.7)
    (On the left,47.9)
    (Viewpoint,22.2)
    (Avg.,40.0)
};
\legend{GS-Reasoner,Ours}
\end{axis}
\end{tikzpicture}
\vspace{-1.5em}
\caption{
Results on three \textbf{Orientation} subsets of
3DSRBench (\textit{In front of}, \textit{On the left}, and
\textit{Viewpoint}). \textit{Avg.} denotes the aggregate performance.
}
\label{fig:downstream_3dsrbench_bar}
\vspace{-2em}
\end{wrapfigure}


A central question is whether orientation grounding provides information that is useful for downstream spatial reasoning.
Importantly, neither model is trained on these downstream benchmarks, allowing us to evaluate whether the learned orientation capability helps downstream tasks without benchmark-specific supervision.
We first evaluate on the \textit{Orientation} subset of 3DSRBench~\citep{ma20253dsrbench}, which is a benchmark designed to assess the 3D spatial reasoning capabilities of VLMs.
As shown in Fig.~\ref{fig:downstream_3dsrbench_bar}, our orientation-grounded model consistently improves over the baseline model (\textit{i.e.,} GS-Reasoner~\citep{chen2025reasoning}), increasing the overall accuracy from 30.0\% to 40.0\%.
These results indicate that, even without 3DSRBench-specific training, strengthening the model's orientation-grounding capability can directly improve spatial reasoning that relies on an object's intrinsic reference frame.

\begin{wraptable}{r}{0.35\textwidth}
\vspace{-2.0em}
\centering
\caption{Downstream situated spatial reasoning results on MSQA.}
\label{tab:downstream_MSQA}
\vspace{0.3em}

\resizebox{0.35\textwidth}{!}{
\begin{tabular}{lcc}
\toprule
\textbf{Model} & \textbf{EM@1 $\uparrow$} & \textbf{EM@1-Strict $\uparrow$} \\
\midrule
GS-Reasoner & 29.02 & 18.05 \\
Ours        & \textbf{32.64} & \textbf{28.18} \\
\bottomrule
\end{tabular}
}
\vspace{-1em}
\end{wraptable}

We further evaluate on MSQA~\citep{linghu2024multi}, which provides a complementary setting for \textbf{situated 3D spatial reasoning}, where directional relations must be interpreted under a spatial reference frame rather than as isolated geometric attributes.
We follow the official evaluation protocol and report Exact Match (EM)@1 and EM@1-Strict under the same benchmark evaluation setting (Table~\ref{tab:downstream_MSQA}).
Although the improvements are less directly tied to object-centric relations than on 3DSRBench, our model still improves both EM@1 and EM@1-Strict, suggesting that directional understanding also transfers to situated spatial reasoning.

Together, these results suggest that orientation grounding serves as a complementary geometric cue for spatial reasoning across diverse downstream reasoning scenarios, particularly when object orientation is part of the spatial context.
By providing a query-conditioned, object-centric frame of reference, it captures information that localization alone cannot recover, especially for symmetric objects and orientation-dependent referring expressions where valid orientations depend on the query.
More downstream evaluation protocols are provided in Appendix~\ref{app:downstream_eval}.

\section{Conclusion}

We introduced \emph{orientation grounding} as a referring 3D grounding task
that extends localization to object-centric orientation and axial symmetry,
and constructed ReferOri, a scalable single- and multi-view benchmark
and training resource.
We show that existing spatial VLMs can
acquire strong orientation-grounding capability through explicit supervision,
structured outputs, and lightweight geometry-aware losses.
Experiments demonstrate substantial gains over orientation-aware VLMs and
object-level orientation foundation models, while downstream evaluations
further show benefits for orientation-related spatial reasoning.
We hope this work encourages more complete spatial grounding beyond object
location alone.
Future work can build richer spatial reasoning datasets upon comprehensive
spatial grounding representations to further study grounding-to-reasoning
transfer. We also discuss the limitations of the current formulation and evaluation in
Appendix~\ref{limitation}.

\clearpage
{\small
\bibliography{refs}
\bibliographystyle{iclr2027_conference}
}

\newpage
\appendix



\section*{AI Use Statement}
Large language models were used only for language polishing and improving
writing clarity. They were not used for research design, experiments, results,
or scientific conclusions. All assisted text was reviewed by the authors.

\section{Limitations}
\label{limitation}
Our annotation pipeline may introduce noise through imperfect segmentation,
reconstruction ambiguity, and coordinate transformation. Manual verification
and an independent annotation audit indicate high annotation consistency,
although residual errors remain, particularly for ambiguous or weakly
directional objects. In addition, a noticeable gap remains between our model
and reconstruction-based object-level orientation estimates, suggesting that
the current VLM has not yet fully absorbed the geometric cues available from
reconstruction.

Although our model performs strongly on multi-view orientation grounding,
dedicated orientation foundation models remain stronger in object-centric
single-view orientation estimation. The two settings are also not directly
equivalent: our model reasons over queried scenes, whereas orientation
foundation models typically operate on cropped objects or masks. Our goal is
therefore not to replace these models, but to extend orientation understanding
from isolated objects to referring grounding in single- and multi-view scenes.
Bridging these complementary capabilities remains an important direction for
future work.

Our model is obtained by fine-tuning an existing 3D VLM, and we do not
comprehensively evaluate whether this adaptation preserves all capabilities of
the base model. Incorporating orientation grounding during pretraining may
provide a more unified way to learn localization, orientation, and reasoning.
Our robustness analyses also reveal residual reliance on upright-scene and
language priors, motivating rotation-aware training, explicit gravity
conditioning, and visually unanswerable examples.

Finally, the current downstream evaluation focuses on orientation-related
spatial reasoning rather than full embodied systems such as manipulation or
navigation. Evaluating whether improved orientation grounding translates to
broader embodied performance remains future work.

\vspace{-1em}
\section{Data Construction Details}
\label{app:data_construction}

This appendix provides additional details for the construction of our orientation grounding data, including the single-view and multi-view pipelines, symmetry estimation, coordinate alignment, annotation-module validation, human verification, and failure cases.

\subsection{Single-View Construction Pipeline}
\label{app:single_view_pipeline}

Our single-view construction pipeline starts from existing referring grounding data with object-level bounding boxes. Given an image $I$ and a referred object box $b$, we first filter out severely blurred images to avoid unstable reconstruction and ambiguous geometry.

We then apply a metric-scale monocular 3D scene reconstruction model to estimate scene depth and camera parameters. This yields a camera-to-world transformation $T_{\text{world}\leftarrow\text{cam}}$ together with a coarse 3D scene representation in a unified metric frame.

Next, we refine the object region using a 2D segmentation model. Specifically, we use the referred box $b$ as input to obtain an object mask $M$, which provides a more precise object boundary than the original bounding box. The masked object, together with the original image, is fed into an object-level 3D reconstruction model, which produces:
\begin{itemize}
    \item an object point cloud $\mathcal{P}$,
    \item an object canonical coordinate system,
    \item and an object-to-camera transformation $T_{\text{cam}\leftarrow\text{obj}}$.
\end{itemize}

We treat the canonical coordinate system of the reconstructed object as an orientation proxy for the object itself. Let $R_{\text{cam}\leftarrow\text{obj}}$ denote the rotation component of $T_{\text{cam}\leftarrow\text{obj}}$, and let $R_{\text{world}\leftarrow\text{cam}}$ denote the rotation component of $T_{\text{world}\leftarrow\text{cam}}$. The object orientation in the world frame is computed as
\[
R_{\text{obj}}^{\text{world}}
=
R_{\text{world}\leftarrow\text{cam}}
\, R_{\text{cam}\leftarrow\text{obj}}.
\]
The final rotation is stored as the ground-truth object orientation in \(SO(3)\), and converted into a continuous 6D representation during training.

Based on the reconstructed point cloud $\mathcal{P}$, we further estimate the object's axial symmetry labels along the $x$, $y$, and $z$ axes. The final annotation tuple for each referred object is
\[
\mathcal{A}=(B,R,S),
\]
where $B$ is the 3D bounding box, $R\in SO(3)$ is the world-frame object orientation, and $S\in\{0,1\}^3$ denotes the symmetry labels.

\subsection{Multi-View Construction Pipeline}
\label{app:multi_view_pipeline}

For multi-view scenes, we exploit view consistency to improve annotation reliability. Given a referred object instance in a multi-view scene, we first identify the two most informative views. In practice, we approximate informativeness by the visible object area and select the top two frames with the largest mask area or bounding-box area.

For each selected frame, we estimate the object orientation independently using the same object-level reconstruction pipeline as in the single-view setting. Each orientation estimate is then transformed into the common scene world frame using the native camera parameters provided by the multi-view scene.

Let the resulting orientations be $R_1^{\text{world}}$ and $R_2^{\text{world}}$. We compute their angular discrepancy
\[
\Delta(R_1,R_2)=\arccos\left(\frac{\mathrm{tr}(R_1 R_2^\top)-1}{2}\right).
\]
If \(\Delta(R_1,R_2) < \tau\), where \(\tau=5^\circ\), we directly accept the annotation. Otherwise, we retain one prediction as a coarse annotation and ask a human annotator to correct it.

This strategy greatly reduces manual effort. Instead of labeling orientation from scratch, annotators only need to adjust a coarse but already plausible initialization.

\subsection{Symmetry Estimation}
\label{app:symmetry_estimation}

We estimate axial symmetry labels directly from the reconstructed point cloud \(\mathcal{P}\). For each canonical axis \(a\in\{x,y,z\}\), we reflect the point cloud across the corresponding symmetry plane and measure the geometric consistency between the reflected point set and the original point cloud. If the discrepancy falls below a predefined threshold, we assign the corresponding axis symmetry label as positive.

Formally, let \(\text{Ref}_a(\mathcal{P})\) denote the reflected point cloud across axis \(a\), and let \(d(\cdot,\cdot)\) denote a bidirectional point-set discrepancy. We mark axis \(a\) as symmetric if
\[
d\big(\mathcal{P}, \text{Ref}_a(\mathcal{P})\big) < \epsilon_a.
\]
This produces the final symmetry label vector
\[
S = (s_x, s_y, s_z), \quad s_a\in\{0,1\}.
\]

In practice, symmetry estimation is reliable for most rigid and structurally regular objects, but can become noisy under severe reconstruction artifacts, heavy occlusion, or near-symmetric shapes.

\subsection{Coordinate Alignment and Sanity Checks}
\label{app:coord_alignment}

Our pipeline combines outputs from scene-level and object-level reconstruction modules, which may adopt different coordinate conventions. To ensure consistency, we explicitly verify:
\begin{itemize}
    \item coordinate convention compatibility across modules,
    \item global axis flips,
    \item handedness mismatch,
    \item and world-frame alignment after transformation.
\end{itemize}

These checks are not conducted only in image space. During verification, we inspect both the original 2D image and the final reconstructed 3D scene with the transformed object orientation. This allows us to detect systematic issues that may not be obvious from a 2D projection alone, such as left-right inversion or incorrect up-axis assignment.

We find that such sanity checks are important in practice, especially when combining independently trained reconstruction modules.

\subsection{Annotation Module Validation}
\label{app:annotation_module}

Since SAM3D~\citep{chen2025sam} provides the object-level orientation proposals
used in our annotation pipeline, we evaluate its orientation estimates on
Ori-COCO as an external reference.
As shown in Table~\ref{tab:ori_coco}, SAM3D achieves 96.3\% accuracy,
compared with 72.4\% and 86.4\% for Orient Anything V1 and V2,
respectively.
This evaluation provides an external sanity check on the orientation proposals
used for large-scale annotation.

\begin{wraptable}{r}{0.4\textwidth}
\centering
\vspace{-2.5em}
\caption{Validation of the annotation module on Ori-COCO (Acc).}
\label{tab:ori_coco}
\resizebox{\linewidth}{!}{
\begin{tabular}{lc}
\toprule
\textbf{Model} & \textbf{Acc} \\
\midrule
Orient Anything V1~\citep{wang2024orient} & 72.4 \\
Orient Anything V2~\citep{wang2026orient} & 86.4 \\
SAM3D~\citep{chen2025sam} & 96.3 \\
\bottomrule
\end{tabular}
}
\vspace{-5em}
\end{wraptable}

\subsection{Human Verification Protocol}
\label{app:human_verification}

For single-view annotations, we evaluate reliability through repeated random
human inspection. We conduct five rounds, each with 100 randomly sampled
examples. During verification, annotators inspect:
\begin{itemize}
    \item the original 2D image,
    \item the target object region, and
    \item the reconstructed 3D scene with the projected object orientation.
\end{itemize}

A sample is counted as correct only when the referred object is correctly
identified and its annotated orientation is visually consistent with both the
source image and reconstructed geometry. Annotators additionally check
coordinate consistency, global axis flips, and handedness mismatches.
Across the five rounds, 96\% of the sampled annotations are judged correct on
average, which we report as the \emph{human-verified correctness rate} rather
than accuracy against ground-truth annotations.

\paragraph{Multi-view annotation protocol.}
For ScanNet multi-view data, where reconstruction is less reliable,
reconstruction outputs are treated only as annotation proposals.
All object annotations are manually inspected.
Annotators check whether the proposed canonical frame and symmetry label are
consistent with the available multi-view evidence and reconstructed geometry.
Incorrect proposals are corrected, while ambiguous or low-confidence cases are
independently re-annotated by a second annotator.
Remaining disagreements are resolved through joint inspection, and samples
without reliable consensus are discarded.
For weakly directional objects, symmetry is assigned only when alternative
directions are genuinely equivalent; if a unique direction exists but cannot
be reliably determined, the annotation is corrected or removed.

\paragraph{Independent audit.}
We further audit 200 images from each of ScanNet, RefCOCO, RefCOCO+, and
RefCOCOg, for a total of 800 images.
Two evaluators independently judge whether each full annotation---including
the referent, orientation frame, and symmetry label---is correct, yielding
1{,}600 binary judgments.
Table~\ref{tab:annotation_audit} reports percentage agreement and Fleiss'
$\kappa$.
Agreement ranges from 96.0\% to 98.8\%, with 97.5\% overall.
The lower $\kappa$ values despite high raw agreement are consistent with the
prevalence effect under the strong imbalance toward annotations judged as
correct.

\begin{table}[h]
\centering
\vspace{-1em}
\caption{Independent annotation audit: agreement between two evaluators.}
\label{tab:annotation_audit}
\begin{tabular}{lcc}
\toprule
\textbf{Dataset} & \textbf{Agreement} & \textbf{Fleiss' $\kappa$} \\
\midrule
ScanNet & 96.00\% & 0.479 \\
RefCOCO & 96.27\% & 0.409 \\
RefCOCO+ & 98.79\% & 0.436 \\
RefCOCOg & 98.68\% & 0.493 \\
\midrule
Overall & 97.45\% & 0.454 \\
\bottomrule
\end{tabular}
\vspace{-2em}
\end{table}

\subsection{Advantages of the Pipeline}
\label{app:pipeline_advantages}

Our construction pipeline has three main advantages.

First, it is \textbf{scalable}. It builds upon existing referring data,
segmentation, and reconstruction modules, allowing orientation annotations to
be generated at large scale without exhaustive manual labeling.

Second, it supports both \textbf{single-view} and \textbf{multi-view}
settings. This enables us to cover both natural-image scenarios and richer 3D
scene understanding benchmarks within a unified formulation.

Third, it is \textbf{extensible}. Since the pipeline is modular, future
advances in scene reconstruction and object reconstruction can be directly
incorporated to improve annotation quality. In this sense, the quality of the
benchmark can continue to improve with stronger underlying 3D perception
models.

\vspace{-0.5em}
\subsection{Failure Cases and Limitations}
\label{app:failure_cases}

Although effective in practice, the pipeline still has several limitations.

\paragraph{Reconstruction ambiguity.}
The canonical coordinate system produced by object-level reconstruction is
used as an orientation proxy, but it is not guaranteed to perfectly match
human-defined intrinsic orientation in all cases. This issue is most pronounced
for highly symmetric objects or objects with weak directional cues.

\paragraph{Segmentation errors.}
Errors in the 2D object mask may propagate to the object reconstruction stage,
which can in turn affect both orientation and symmetry estimation.

\paragraph{Occlusion and incomplete geometry.}
Heavy occlusion may lead to incomplete or distorted reconstructions, making
symmetry estimation and orientation inference less reliable.

\paragraph{Cross-module noise.}
The final annotation quality depends on the consistency of scene reconstruction,
object reconstruction, and coordinate transformation. Although we explicitly
check for major mismatches, residual noise may still remain.

We therefore regard the constructed annotations as scalable supervision with
bounded noise, rather than perfectly noise-free labels.

\vspace{-1em}
\section{Evaluation Protocol Details}
\label{app:evaluation_details}

\subsection{Unified Rotation Conversion}
\label{app:rotation_conversion}

All continuous-orientation benchmarks are evaluated after converting each
method's native output into a $3\times3$ rotation matrix in the same
right-handed reference frame ($x$ right, $y$ forward, $z$ up):
\begin{itemize}
    \item \textbf{SpatialReasoner}~\citep{ma2025spatialreasoner} predicts a
    front direction under an assumed world-up direction; we complete it to a
    full rotation matrix and map it into the shared frame.
    
    \item \textbf{VG-LLM}~\citep{zheng2025learning} predicts Euler angles; we
    compose them into a rotation matrix following its published convention.
    
    \item \textbf{Orient Anything V1/V2}~\citep{wang2024orient,wang2026orient}
    predict azimuth--polar--rotation angles; we convert them into rotation
    matrices following the Orient Anything convention.
    
    \item \textbf{OG-VLM} predicts a continuous 6D rotation representation;
    we orthogonalize it via Gram--Schmidt and map it into the same reference
    frame.
\end{itemize}

After conversion, all methods are evaluated using the same $SO(3)$ geodesic
error, so differences in native orientation parameterization do not introduce
representation-specific evaluation metrics.

\subsection{Symmetry-Aware Evaluation}
\label{app:symmetry_treatment}

The symmetry-aware evaluation in Sec.~\ref{sec:task_formulation} uses the
model-predicted axial symmetry labels rather than ground-truth symmetry.
For OG-VLM, the predicted labels
$\hat{S}=(\hat{s}_x,\hat{s}_y,\hat{s}_z)$ determine the set of
symmetry-equivalent transformations $\mathcal{G}(\hat{S})$, and the orientation
error is computed as the minimum $SO(3)$ geodesic distance over this set.

For baselines that do not provide compatible axial symmetry predictions,
including SpatialReasoner and VG-LLM, we use the identity equivalence set
$\mathcal{G}=\{I\}$, corresponding to no symmetry adjustment.
Thus, ground-truth symmetry labels are never used to relax orientation errors
during evaluation.

Our axis-specific symmetry representation differs from the
rotational-order-$k$ representation used by Orient Anything
V2~\citep{wang2026orient}. As shown in
Table~\ref{tab:metric_decomposition}, symmetry adjustment produces
dataset-dependent improvements, while the relatively modest changes on most
benchmarks indicate that the reported orientation performance is not primarily
driven by symmetry handling.
\subsection{Benchmark Splits and Metrics}
\label{app:eval_protocol}

RefCOCO/RefCOCO+/RefCOCOg, ScanRefer, and Multi3DRefer are constructed by converting the test splits of the corresponding referring datasets into orientation-grounding test sets while preserving their original referring structure. Ori-COCO is adopted from Orient Anything~\citep{wang2024orient}. For ARKitScenes~\citep{baruch2021arkitscenes}, which already provides oriented 3D bounding-box annotations, we directly use its validation split following common practice.

For all continuous benchmarks, every method's native output is converted into a rotation matrix in the same right-handed reference frame before evaluation. Specifically, we convert Euler angles for VG-LLM, a front direction under an assumed world-up for SpatialReasoner, azimuth--polar--rotation angles for Orient Anything, and the continuous 6D rotation representation used by our method. All predictions are then evaluated using the same symmetry-aware \(SO(3)\) geodesic metric.

Ori-COCO is treated separately because it does not provide fine-grained \(SO(3)\) annotations; instead, its labels partition the horizontal orientation space into eight \(45^\circ\) sectors. Following its native protocol, we bin each predicted orientation into one of these sectors and report classification accuracy. Using Acc@15/Acc@30 on this benchmark would primarily reflect label granularity rather than method quality. Accordingly, Ori-COCO results in Table~\ref{tab:orientation_results} are reported under its native metric. Table~\ref{tab:ori_coco} uses Ori-COCO to validate SAM3D as our annotation module under the original Orient Anything protocol, whereas Table~\ref{tab:orientation_results} uses it as an independently annotated external benchmark.

\subsection{Downstream Evaluation Protocol}
\label{app:downstream_eval}

\paragraph{3DSRBench.}
For 3DSRBench~\citep{ma20253dsrbench}, we evaluate the \textbf{Orientation} subset, including \emph{In front of}, \emph{On the left}, and \emph{Viewpoint}, following the benchmark's official evaluation protocol. The benchmark uses CircularEval, which evaluates each question under multiple answer-choice permutations and counts it as correct only when the model remains correct across the permutations. Its test set also includes FlipEval-augmented examples to reduce left/right response bias. We report accuracy for each relation type and the aggregate accuracy over the Orientation subset.

\paragraph{MSQA.}
For MSQA~\citep{linghu2024multi}, we follow the MSR3D evaluation protocol and report \textbf{EM@1} and \textbf{EM@1-Strict}. Predictions and reference answers are normalized by lowercasing, whitespace normalization, punctuation removal, spelling correction, number-word conversion, and article removal where applicable. EM@1-Strict requires an exact match with at least one normalized reference answer, whereas EM@1 additionally accepts containment matches between the normalized prediction and reference answers. Empty predictions receive zero under both metrics, and dataset-level scores are averaged over all evaluated questions.

Neither model receives benchmark-specific QA supervision or task-specific fine-tuning on 3DSRBench or MSQA.

\vspace{-1em}
\section{Additional Results and Analyses}
\label{app:additional_analyses}

\subsection{Sensitivity to Image-Plane Rotation}
\begin{wraptable}{r}{0.72\textwidth}
\vspace{-2em}
\centering
\caption{Sensitivity of OG-VLM to image-plane rotation of the input view.
Axis-wise and overall mean orientation errors are in degrees; Acc@15/Acc@30
are reported in \%.}
\label{tab:rotation_robustness}
\vspace{0.3em}
\resizebox{0.72\textwidth}{!}{
\begin{tabular}{lcccccc}
\toprule
\textbf{Rotation} & \textbf{X Err.} & \textbf{Y Err.} & \textbf{Z Err.} &
\textbf{Mean Err.} & \textbf{Acc@15} & \textbf{Acc@30} \\
\midrule
$90^\circ$  & 69.0 & 51.3 & 83.9 & 60.2 & 0.05 & 0.20 \\
$180^\circ$ & 40.1 & 44.5 & 18.7 & 42.3 & 16.71 & 37.32 \\
$270^\circ$ & 68.4 & 52.1 & 83.6 & 60.3 & 0.05 & 0.25 \\
\bottomrule
\end{tabular}}
\vspace{-1em}
\end{wraptable}

\label{app:rotation_robustness}
We evaluate OG-VLM on inputs rotated in the image plane by $90^\circ$, $180^\circ$, and $270^\circ$. As shown in Table~\ref{tab:rotation_robustness}, $90^\circ$ and $270^\circ$ rotations break the alignment between image-top and world-up and increase the Z-axis error to roughly $84^\circ$, collapsing Acc@15; a $180^\circ$ rotation preserves the vertical axis and mainly increases horizontal facing errors. This reveals a reliance on upright-scene priors; rotation-aware augmentation or explicit gravity conditioning are natural remedies, which we leave to future.

\subsection{Input Frames Ablation}
\label{app:views}

We study how the number of input frames affects orientation grounding on Multi3DRefer. As shown in Table~\ref{tab:multi3drefer_ori}, performance drops from multi-view to single-view, and further to language-only input, showing that orientation grounding depends strongly on visual and geometric evidence. Multi-view input provides complementary cues for resolving occlusion and structural ambiguity, while single-view input offers only limited local appearance and shape information.
\begin{wraptable}{r}{0.46\textwidth}
\vspace{-1em}
\centering
\caption{Results on Multi3DRefer$^{*}$ under different settings.}
\label{tab:multi3drefer_ori}
\resizebox{0.46\textwidth}{!}{
\begin{tabular}{llcc}
\toprule
\textbf{Model} & \textbf{Settings} & \textbf{Acc@15} & \textbf{Acc@30} \\
\midrule
\multirow{3}{*}{Ours (only data)}
& multi-view  & 68.08 & 70.64 \\
& single-view & 58.43 & 61.46 \\
& language    & 20.00 & 40.00 \\
\midrule
\multirow{3}{*}{Ours (+special token)}
& multi-view  & 70.50 & 73.06 \\
& single-view & 56.88 & 60.33 \\
& language    & 8.04 & 10.80 \\
\midrule
\multirow{3}{*}{Ours (+geometric loss)}
& multi-view  & 71.50 & 74.20 \\
& single-view & 64.25 & 66.92 \\
& language    & 22.22 & 22.22 \\
\bottomrule
\end{tabular}
}
\vspace{-1.5em}
\end{wraptable}


Notably, the model still outputs orientation predictions even without visual input. Such non-zero performance is expected for pretrained VLMs, which retain category-level and scene-level language priors. Our goal in this ablation is therefore not to eliminate these priors, but to quantify the additional benefit of visual evidence, particularly multi-view observations, for orientation estimation. The language-only performance mainly reflects residual priors rather than genuine grounding ability; while they may support coarse guesses from object categories or common scene configurations, they are insufficient for robust orientation grounding in ambiguous or non-canonical cases.

\subsection{Downstream Spatial Reasoning Analysis}
\label{app:downstream_analysis}
The three orientation relations in 3DSRBench probe complementary uses of the grounded object-centric frame. \emph{In front of} primarily depends on the intrinsic front axis, while \emph{On the left} depends on the intrinsic right/left axis. \emph{Viewpoint} additionally requires relating the object's intrinsic frame to the camera frame, making it a more challenging reference-frame transformation. The consistent gains across all three relations therefore suggest that improved orientation grounding benefits multiple forms of orientation-dependent spatial reasoning rather than a single relation type.

\subsection{Localization Results}
\label{app:localization}
\begin{table}[t]
\centering
\vspace{-1em}
\caption{Localization grounding results at Acc25\% on ScanRefer, Multi3DRefer, RefCOCO, RefCOCO+, and RefCOCOg.}
\label{tab:acc25_results}
\begin{tabular}{lccccc}
\toprule
\textbf{Model} & \textbf{ScanRefer} & \textbf{Multi3DRefer} & \textbf{RefCOCO} & \textbf{RefCOCO+} & \textbf{RefCOCOg} \\
& \textbf{Acc 25\%} & \textbf{Acc 25\%} & \textbf{Acc 25\%} & \textbf{Acc 25\%} & \textbf{Acc 25\%} \\
\midrule
Ours (only data)         & 55.68 & 55.43 & 71.54 & 68.91 & 62.37 \\
Ours (+special token)    & 58.95 & 57.38 & 71.84 & 69.26 & 62.42 \\
Ours (+geometric loss)   & \textbf{61.37} & \textbf{60.08} & \textbf{72.67} & \textbf{70.31} & \textbf{63.38} \\
\bottomrule
\end{tabular}
\end{table}
We additionally report localization grounding results to examine whether adding explicit orientation supervision preserves the base model's localization capability. These results complement the orientation-only evaluation in the main paper by showing the behavior of the localization component under the same training variants.

\vspace{-1em}
\subsection{Qualitative Comparison with Baselines}
\begin{figure*}[t]
\vspace{-1em}
    \centering
    \includegraphics[width=1\linewidth]{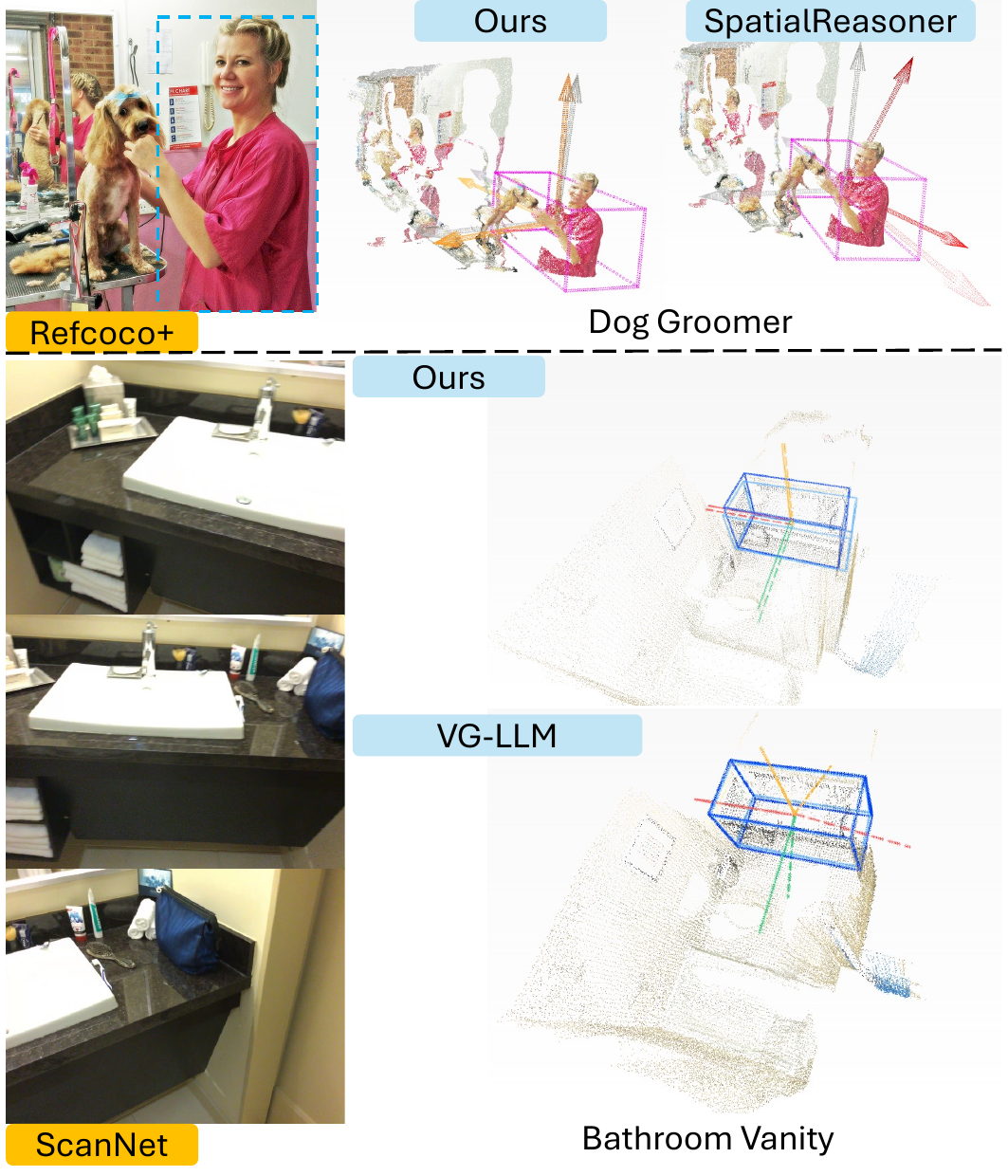}
    \vspace{-2em}
    \caption{
Qualitative orientation-grounding results on RefCOCO+ and ScanNet.
For RefCOCO+, gray arrows denote ground truth, while orange and red arrows denote predictions from Ours and SpatialReasoner, respectively.
For ScanNet, solid axes denote ground truth and dashed axes denote predictions from Ours and VG-LLM, respectively.
}
\vspace{-1em}
    \label{fig:qualitative_results}
\end{figure*}

Figure~\ref{fig:qualitative_results} shows that our predictions better align
with ground-truth orientations than SpatialReasoner on RefCOCO+ and VG-LLM on
ScanNet across both single-view and multi-view settings.

\label{app:qualitative}

\newpage
\section{Training and Geometry-Aware Loss Details}
\label{app:geo_loss_details}

This appendix provides the detailed formulation of the geometry-aware losses used in Sec.~\ref{sec:geo_loss}, including the box loss, orientation loss, symmetry loss, and the implementation details of the corresponding prediction heads.

\subsection{Dataset Construction and Training Details}
\label{app:training_details}

For multi-view scenes, we construct orientation-grounding data from Locate3D~\citep{arnaud2025locate}, SR3D~\citep{zhang2025sr3d}, NR3D~\citep{achlioptas2020referit3d}, ScanRefer~\citep{chen2020scanrefer}, and Multi3DRefer~\citep{zhang2023multi3drefer}. For single-view scenes, we use RefCOCO~\citep{yu2016modeling}. For each referred object, we construct a language query from the original referring expression and a box query from the corresponding bounding box. The resulting dataset contains 331K multi-view queries and 387K single-view queries, with the two query types approximately balanced at a 1:1 ratio.

We use GS-Reasoner~\citep{chen2025reasoning} as the base model. Training is performed on 4$\times$H200 GPUs, with each full training run taking approximately 120 hours. Loss coefficients and additional optimization details are summarized in Sec.~\ref{app:geo_hyper}.

\subsection{Prediction Heads}
\label{app:pred_heads}

To explicitly disentangle localization and orientation learning, we adopt multiple lightweight prediction heads on top of the hidden states, following the spirit of VGR~\citep{wang2025vgr}, which shows that geometric supervision can be effectively introduced through small MLP heads over language model representations.

Concretely, we use:
\begin{itemize}
    \item a \textbf{box head} on the hidden state at the \texttt{</bbox>} token to regress a 3D box,
    \item an \textbf{orientation head} on the hidden state at the \texttt{</ori>} token to regress the continuous 6D orientation representation,
    \item and a \textbf{symmetry head} to predict symmetry-related geometry.
\end{itemize}
This decoupling reduces interference between localization and orientation prediction, while allowing each sub-task to receive task-specific geometric supervision.

\subsection{Detailed Box Loss}
\label{app:box_loss}

For 3D grounding, the model predicts an axis-aligned box
\begin{equation}
B=[x_{\min},y_{\min},z_{\min},x_{\max},y_{\max},z_{\max}] \in \mathbb{R}^6.
\end{equation}
Let \(B_p\) and \(B_g\) denote the predicted and ground-truth boxes, respectively. The box loss in Eq.~\eqref{eq:box_loss} consists of three terms:
\begin{equation}
\mathcal{L}_{\text{box}}
=
\alpha_{\text{giou}}\mathcal{L}_{\text{GIoU}}
+
\alpha_{\text{scale}}\mathcal{L}_{\text{scale}}
+
\alpha_{\text{center}}\mathcal{L}_{\text{center}}.
\end{equation}

\paragraph{GIoU loss.}
We first define the 3D IoU as
\begin{equation}
\mathrm{IoU}(B_p,B_g)=\frac{|B_p\cap B_g|}{|B_p\cup B_g|}.
\end{equation}
Let \(C\) denote the smallest enclosing box covering both \(B_p\) and \(B_g\). The generalized IoU is
\begin{equation}
\mathrm{GIoU}(B_p,B_g)=
\mathrm{IoU}(B_p,B_g)
-
\frac{|C|-|B_p\cup B_g|}{|C|},
\end{equation}
and the corresponding loss is
\begin{equation}
\mathcal{L}_{\text{GIoU}} = 1-\mathrm{GIoU}(B_p,B_g).
\end{equation}

\paragraph{Scale loss.}
Let
\begin{equation}
\mathbf{s}_p=\mathbf{b}^{p}_{\max}-\mathbf{b}^{p}_{\min},\qquad
\mathbf{s}_g=\mathbf{b}^{g}_{\max}-\mathbf{b}^{g}_{\min}
\end{equation}
denote the predicted and ground-truth box sizes. We measure the relative scale discrepancy along each axis as
\begin{equation}
\mathcal{L}_{\text{scale}}
=
\frac{1}{3}\sum_{k\in\{x,y,z\}}
\frac{|s_{p,k}-s_{g,k}|}{\max(s_{p,k},s_{g,k})+\epsilon}.
\end{equation}

\paragraph{Center loss.}
Let
\begin{equation}
\mathbf{c}_p=\frac{\mathbf{b}^{p}_{\min}+\mathbf{b}^{p}_{\max}}{2},\qquad
\mathbf{c}_g=\frac{\mathbf{b}^{g}_{\min}+\mathbf{b}^{g}_{\max}}{2}
\end{equation}
denote the predicted and ground-truth box centers, and let
\begin{equation}
d_g=\|\mathbf{b}^{g}_{\max}-\mathbf{b}^{g}_{\min}\|_2
\end{equation}
be the diagonal length of the ground-truth box. We define the normalized center loss as
\begin{equation}
\mathcal{L}_{\text{center}}
=
\min\left(
\frac{\|\mathbf{c}_p-\mathbf{c}_g\|_2}{d_g+\epsilon},
1
\right).
\end{equation}

\subsection{Detailed Orientation Loss}
\label{app:ori_loss}

The ground-truth orientation is represented as a rotation \(R\in SO(3)\), while training adopts its continuous 6D representation. From the hidden state at the \texttt{</ori>} token, the orientation head regresses two 3D vectors
\begin{equation}
\mathbf{a}_1,\mathbf{a}_2 \in \mathbb{R}^3.
\end{equation}
We then apply Gram--Schmidt orthogonalization to recover two orthonormal axes \(\hat{\mathbf r}\) and \(\hat{\mathbf f}\), corresponding to the predicted right and front directions.

Let \(\mathbf r,\mathbf f\) denote the ground-truth right and front axes. As defined in Eq.~\eqref{eq:ori_loss}, the orientation loss is
\begin{equation}
\mathcal{L}_{\text{ori}}
=
\beta_{\text{align}}\mathcal{L}_{\text{align}}
+
\beta_{\text{ortho}}\mathcal{L}_{\text{ortho}}.
\end{equation}

\paragraph{Axis alignment loss.}
We define
\begin{equation}
\mathcal{L}_{\text{align}}
=
\frac{1}{2}
\left[
\big(1-\cos(\hat{\mathbf r},\mathbf r)+\epsilon\big)^{\lambda}
+
\big(1-\cos(\hat{\mathbf f},\mathbf f)+\epsilon\big)^{\lambda}
\right].
\label{eq:app_align_loss}
\end{equation}
Compared with a plain cosine loss, the exponent \(\lambda\) allows us to adapt the sensitivity of the alignment term during training. When the alignment error becomes sufficiently small, decreasing \(\lambda\) prevents the loss from saturating too early and preserves optimization sensitivity in the low-error regime.

\paragraph{Orthogonality loss.}
To ensure that the predicted axes remain geometrically valid, we use
\begin{equation}
\mathcal{L}_{\text{ortho}} = |\cos(\hat{\mathbf r},\hat{\mathbf f})|.
\end{equation}
This term penalizes non-orthogonal predictions and stabilizes the recovered orientation basis.

\subsection{Adaptive Exponent Schedule}
\label{app:lambda_schedule}

In practice, the exponent \(\lambda\) in Eq.~\eqref{eq:app_align_loss} is adaptively reduced during training:
\begin{equation}
\lambda =
\begin{cases}
1, & \mathcal{L}_{\text{align}} > \tau_1,\\
1/2, & \tau_2 < \mathcal{L}_{\text{align}} \le \tau_1,\\
1/4, & \mathcal{L}_{\text{align}} \le \tau_2,
\end{cases}
\end{equation}
where \(\tau_1\) and \(\tau_2\) are predefined thresholds. Intuitively, this strategy makes the optimization initially focus on coarse orientation alignment, and later remain sensitive to fine-grained directional refinement.

\subsection{Detailed Symmetry Loss}
\label{app:sym_loss}

To preserve axis-level structural regularity, we further supervise the predicted symmetry representation. Let \(\hat{\mathbf s}\) and \(\mathbf s\) denote the predicted and ground-truth symmetry vectors, respectively. We use a cosine-based loss:
\begin{equation}
\mathcal{L}_{\text{sym}} = 1-\cos(\hat{\mathbf s},\mathbf s).
\end{equation}

This loss complements the token-level symmetry prediction by encouraging the model to preserve geometric symmetry structure in a continuous feature space. In this sense, token-level decoding captures the discrete output format, while the auxiliary symmetry loss provides additional geometric regularization.

\subsection{Hyperparameters and Implementation Details}
\label{app:geo_hyper}
The default coefficients are set as
\[
\alpha_{\text{giou}}=0.5,\quad
\alpha_{\text{scale}}=0.25,\quad
\alpha_{\text{center}}=0.25,\quad
\beta_{\text{align}}=0.8,\quad
\beta_{\text{ortho}}=0.2.
\]
For the overall objective, \(\lambda_{\text{box}}, \lambda_{\text{ori}}, \lambda_{\text{sym}}\) are linearly warmed up during training. In addition, after step 2000, span-aware weighting is applied to the language modeling loss, with weight \(3\) for tokens inside the \texttt{<bbox>} and \texttt{<ori>} spans and weight \(1\) elsewhere.

\ifdefined\includeNeurIPSChecklist
  \newpage
  \section*{NeurIPS Paper Checklist}

\begin{enumerate}

\item {\bf Claims}
    \item[] Question: Do the main claims made in the abstract and introduction accurately reflect the paper's contributions and scope?
    \item[] Answer:\answerYes{} 
    \item[] Justification: The main claims made in the abstract and introduction accurately reflect the paper's contributions and scope.

\item {\bf Limitations}
    \item[] Question: Does the paper discuss the limitations of the work performed by the authors?
    \item[] Answer: \answerYes{} 
    \item[] Justification: We provide limitation discussion in Appendix~\ref{limitation}

\item {\bf Theory assumptions and proofs}
    \item[] Question: For each theoretical result, does the paper provide the full set of assumptions and a complete (and correct) proof?
    \item[] Answer: \answerNA{} 
    \item[] Justification: The paper does not include theoretical results.

    \item {\bf Experimental result reproducibility}
    \item[] Question: Does the paper fully disclose all the information needed to reproduce the main experimental results of the paper to the extent that it affects the main claims and/or conclusions of the paper (regardless of whether the code and data are provided or not)?
    \item[] Answer: \answerYes{} 
    \item[] Justification: All the implementation details are provided in main paper and appendix.

\item {\bf Open access to data and code}
    \item[] Question: Does the paper provide open access to the data and code, with sufficient instructions to faithfully reproduce the main experimental results, as described in supplemental material?
    \item[] Answer: \answerNo{} 
    \item[] Justification:  The full code repository will be released upon the acceptance of the paper

\item {\bf Experimental setting/details}
    \item[] Question: Does the paper specify all the training and test details (e.g., data splits, hyperparameters, how they were chosen, type of optimizer) necessary to understand the results?
    \item[] Answer: \answerYes{} 
    \item[] Justification: The paper specify all the training and test details.

\item {\bf Experiment statistical significance}
    \item[] Question: Does the paper report error bars suitably and correctly defined or other appropriate information about the statistical significance of the experiments?
    \item[] Answer: \answerNo{} 
    \item[] Justification: Error bars are not reported because it would be too computationally expensive. And the expriments are conducted on large scale dataset to provides a reliable estimate of overall performance

\item {\bf Experiments compute resources}
    \item[] Question: For each experiment, does the paper provide sufficient information on the computer resources (type of compute workers, memory, time of execution) needed to reproduce the experiments?
    \item[] Answer:  \answerYes{} 
    \item[] Justification: The paper provides sufficient information on the computer resources.683
    
\item {\bf Code of ethics}
    \item[] Question: Does the research conducted in the paper conform, in every respect, with the NeurIPS Code of Ethics \url{https://neurips.cc/public/EthicsGuidelines}?
    \item[] Answer: \answerYes{} 
    \item[] Justification: The research conducted in the paper conform, in every respect, with the NeurIPS Code of Ethics.

\item {\bf Broader impacts}
    \item[] Question: Does the paper discuss both potential positive societal impacts and negative societal impacts of the work performed?
    \item[] Answer:  \answerNA{} 
    \item[] Justification: There is no societal impact of the work performed.
    
\item {\bf Safeguards}
    \item[] Question: Does the paper describe safeguards that have been put in place for responsible release of data or models that have a high risk for misuse (e.g., pre-trained language models, image generators, or scraped datasets)?
    \item[] Answer:  \answerNA{} 
    \item[] Justification: The paper poses no such risks.

\item {\bf Licenses for existing assets}
    \item[] Question: Are the creators or original owners of assets (e.g., code, data, models), used in the paper, properly credited and are the license and terms of use explicitly mentioned and properly respected?
    \item[] Answer: \answerYes{} 
    \item[] Justification: The original owners of assets are cited

\item {\bf New assets}
    \item[] Question: Are new assets introduced in the paper well documented and is the documentation provided alongside the assets?
    \item[] Answer: \answerNA{} 
    \item[] Justification: No new assets introduced.

\item {\bf Crowdsourcing and research with human subjects}
    \item[] Question: For crowdsourcing experiments and research with human subjects, does the paper include the full text of instructions given to participants and screenshots, if applicable, as well as details about compensation (if any)? 
    \item[] Answer: \answerNA{}{} 
    \item[] Justification: The paper does not involve crowdsourcing nor research with human subjects.

\item {\bf Institutional review board (IRB) approvals or equivalent for research with human subjects}
    \item[] Question: Does the paper describe potential risks incurred by study participants, whether such risks were disclosed to the subjects, and whether Institutional Review Board (IRB) approvals (or an equivalent approval/review based on the requirements of your country or institution) were obtained?
    \item[] Answer: \answerNA{} 
    \item[] Justification: The paper does not involve crowdsourcing nor research with human subjects.
\item {\bf Declaration of LLM usage}
    \item[] Question: Does the paper describe the usage of LLMs if it is an important, original, or non-standard component of the core methods in this research? Note that if the LLM is used only for writing, editing, or formatting purposes and does \emph{not} impact the core methodology, scientific rigor, or originality of the research, declaration is not required.
    \item[] Answer: \answerNA{} 
    \item[] Justification: The core method development in this research does not involve LLMs.
\end{enumerate}
\fi

\end{document}